\documentclass[10pt,twoside,a4paper]{IEEEtran}

\IEEEoverridecommandlockouts
\usepackage{cite}
\usepackage{amsmath,amssymb,amsfonts}
\usepackage{algorithmic}
\usepackage{graphicx}
\usepackage{textcomp}
\usepackage[ruled,vlined,noresetcount]{algorithm2e} 
\usepackage{amsmath}
\usepackage{float}

\usepackage[a4paper,left=1.35cm,right=1.35cm,top=2cm,bottom=4.2cm]{geometry}

\def\BibTeX{{\rm B\kern-.05em{\sc i\kern-.025em b}\kern-.08em
    T\kern-.1667em\lower.7ex\hbox{E}\kern-.125emX}}

\begin{document}

\title{Hybrid Vision Transformer-Mamba Framework for Autism Diagnosis via Eye-Tracking Analysis}

\author{\IEEEauthorblockN{
Wafaa Kasri\IEEEauthorrefmark{1}, 
Yassine Himeur\IEEEauthorrefmark{2}, 
Abigail Copiaco\IEEEauthorrefmark{2},
Wathiq Mansoor\IEEEauthorrefmark{2}
Ammar Albanna\IEEEauthorrefmark{3} 
Valsamma Eapen\IEEEauthorrefmark{4}
}\\
\IEEEauthorblockA{\IEEEauthorrefmark{1}
Faculty of Science and Technology, Tissemsilt University, Bougara~38000, Algeria}\\
\IEEEauthorblockA{\IEEEauthorrefmark{2}College of Engineering and Information Technology University of Dubai Dubai UAE (yhimeur@ud.ac.ae)}\\
\IEEEauthorblockA{\IEEEauthorrefmark{3}College of Medicine and Health Sciences, Mohammed Bin Rashid University Dubai, UAE}\\
\IEEEauthorblockA{\IEEEauthorrefmark{4}School of Clinical Medicine
University of New South Wales, Australia (v.eapen@unsw.edu.au)}\\
}

\maketitle
\thispagestyle{empty}
\pagestyle{empty}

\begin{abstract}
Accurate ASD diagnosis is vital for early intervention. This study presents a hybrid deep learning framework combining Vision Transformers (ViT) and Vision Mamba to detect Autism Spectrum Disorder (ASD) using eye-tracking data. The model uses attention-based fusion to integrate visual, speech, and facial cues, capturing both spatial and temporal dynamics. Unlike traditional handcrafted methods, it applies state-of-the-art deep learning and explainable AI techniques to enhance diagnostic accuracy and transparency. Tested on the Saliency4ASD dataset, the proposed ViT-Mamba model outperformed existing methods, achieving 0.96 accuracy, 0.95 F1-score, 0.97 sensitivity, and 0.94 specificity. These findings show the model’s promise for scalable, interpretable ASD screening, especially in resource-constrained or remote clinical settings where access to expert diagnosis is limited.

\end{abstract}


\IEEEpeerreviewmaketitle

\section{Introduction}
Autism Spectrum Disorder (ASD) is a multifaceted neurodevelopmental condition marked by difficulties in social interaction, repetitive behaviors, and heightened sensory sensitivities \cite{chen2024deep}. Timely and precise diagnosis is essential for initiating effective interventions; however, conventional methods largely depend on subjective evaluations. These assessments are not only resource-intensive—requiring time, cost, and specialized expertise—but also prone to variability, often delaying intervention and impacting developmental outcomes \cite{qi2024visual}. With the global incidence of ASD steadily increasing, the demand for more accurate, scalable, and accessible diagnostic solutions is becoming increasingly critical \cite{mumenin2024asdnet}.

In recent years, eye-tracking technology has gained recognition as a valuable tool in ASD detection, offering objective, quantifiable insights into individuals’ visual attention \cite{wei2024early}. People on the autism spectrum frequently display distinctive gaze behaviors—for example, spending less time looking at human faces or showing irregular eye movement patterns when processing social cues. Such gaze-based differences can be leveraged to train diagnostic models capable of distinguishing ASD from typical development with notable precision \cite{alsaidi2024convolutional}. Yet, despite the promise of this approach, current eye-tracking models are hindered by several challenges, including limited dataset diversity, inconsistent feature extraction methods, and reduced generalizability across varied populations \cite{cheekaty2024enhanced}.

One notable effort to incorporate eye-tracking tech into ASD diagnosis is EyeTism \cite{duan2019dataset}, a model created to analyse gaze-based features for detecting autism. Though EyeTism has shown some promising outcomes, it carries several drawbacks—like its dependance on hand-crafted features, limited multi-modal fusion, and poor interpretability. Also, many current models face data inefficiencies and don't really tap into the full potential of modern deep learning, which may result in biases or reduced performance in real-world clinical use. These gaps highlight the need for more sophisticated frameworks that make use of state-of-the-art AI to boost diagnostic reliability and accuracy.

In this paper, we present a new hybrid model that combines ViT with Vision Mamba for improving ASD diagnosis using eye-tracking data. The core contributions include: (i) building a ViT-Mamba model that captures both spatial fixation maps and long-range visual attention over time; (ii) improving the Saliency4ASD dataset by adding more varied and enriched gaze samples to boost generalizability; (iii) applying advanced feature extraction methods to uncover meaningful spatiotemporal gaze patterns linked with ASD; (iv) integrating multiple data types—like facial expressions, eye movement, and speech—via an attention fusion strategy for stronger diagnostics; (v) leveraging cutting-edge deep learning tools for high-accuracy classification tasks; and (vi) embedding explainability layers to make the model’s predictions more interpretable. Together, these additions aim to move ASD screening forward by delivering tools that are practical, scalable, and clinically insightful.

\section{Related Work}

Early ASD diagnosis has increasingly leaned on eye-tracking (ET) data combined with machine learning (ML) models to deliver more scalable and objective screening tools. Several works have managed to turn ET scanpaths into visual features for classification, with neural networks achieving strong results (AUC $>$ 0.9) \cite{Carette2019ASD}. Systematic reviews also underline the value of deep learning (DL) models—especially convolutional neural networks (CNNs) and generative adversarial networks (GANs)—in ASD-related neuroimaging, though ethical issues around transparency and consent are still not fully settled \cite{Halkiopoulos2023EmotionDetection}.

Various ML and DL methods have been applied to ASD detection. T-CNN-ASD achieved around 95.59\% accuracy \cite{Alsaidi2023ASDPrediction}; CNN-GRU-ANN combinations modelled gaze sequences effectively \cite{Benabderrahmane2024}; and hierarchical support vector machines (SVMs) reached up to 94.28\% accuracy \cite{Xia2020}. Notably, CNN-RNN-based scanpath models went even higher, up to 97\% accuracy \cite{Benabderrahmane2024}. Other hybrid models, like GoogleNet plus SVM, scored 95.5\% \cite{Jeyarani2024}, while BiLSTM, GRU, and CNN-LSTM architectures have peaked at 98.33\% \cite{Ahmed2024}. Still, issues like generalisability and how well the models can be explained remain ongoing concerns \cite{Ahmed2024Early}.

To fill those gaps, newer work is turning to transformer-based models. Vision Transformers (ViTs) \cite{cao2023vitasd} have changed the game in computer vision by using self-attention to model global spatial features. They’ve shown strong performance in clinical contexts like tumour detection, organ segmentation, and pathology imaging—where their holistic feature learning beats out traditional CNN-based systems.

Alongside ViTs, the Vision Mamba architecure \cite{liu2024visionmamba} is gaining ground as a promising state-space model that excels at modeling long-range temporal sequences. Initially introduced for sequential signals like ECG and EEG, Vision Mamba brings efficient computation and low memory usage—making it well-suited for processing time-series medical data. Thanks to its state-space design, it often outperforms standard RNNs and LSTMs in tracking nuanced, time-dependent fluctuations that are key for early diagnosis.

Table~\ref{tab:traditional_vs_dl_asd} draws a practical comparison between older machine learning (ML) approaches and newer deep learning (DL) techniques for ASD screening using eye-tracking data. Traditional models like Random Forest (RF) \cite{salhofer2023faces}, XGBoost \cite{hameed2023diagnose,anonymous2023biomarkers}, and Support Vector Classifier (SVC) \cite{ahmed2024early1} are appreciated for their interpretability and decent performance on low-dimensional structured data. However, these methods tend to fall short when faced with the complex, high-dimensional nature of gaze sequences.

By contrast, DL models such as ViT \cite{qasem2024deep,lu2022review}, CNN-LSTM \cite{zhou2024multimodal,alenezi2024multilevel}, and the newer Mamba model are better equipped to learn intricate spatiotemporal patterns from raw data—no handcrafted features needed. Though they require more compute, their capacity to handle multiple data types and extract subtle behavioral signals makes them a strong fit for robust and scalable ASD diagnosis systems.

\begin{table*}[t]
\centering
\scriptsize
\caption{Analysis and comparison of some existing ASD Diagnosis frameworks. }
\begin{tabular}{p{2.8cm}|p{1.4cm}|p{4.2cm}|p{3.8cm}|p{4.2cm}}
\hline
\textbf{Model Type} & \textbf{Category} & \textbf{Strengths} & \textbf{Weaknesses} & \textbf{Best Use Case in ASD Context} \\
\hline
Random Forest (RF) \cite{salhofer2023faces} & Traditional ML & Easy to interpret, good for small datasets & Struggles with high-dimensional visual features & Initial screening using tabular eye-tracking metrics \\
\hline
XGBoost \cite{hameed2023diagnose,anonymous2023biomarkers} & Traditional ML & High accuracy, handles non-linear data well & Requires careful tuning, less transparent & Boosted classification on structured visual features \\
\hline
Support Vector Classifier (SVC) \cite{ahmed2024early1} & Traditional ML & Good for binary classification, effective with clear margins & Not scalable for large, noisy datasets & Binary risk classification on engineered features \\
\hline
ViT \cite{qasem2024deep,lu2022review} & Deep Learning & Excellent at modeling global features in images & Computationally expensive, needs large data & High-dimensional eye image sequence classification \\
\hline
CNN-LSTM \cite{zhou2024multimodal,alenezi2024multilevel}& Deep Learning & Captures both spatial and temporal dependencies & Complex architecture, harder to train & Gaze trajectory classification over time \\
\hline
Mamba [Proposed] & Deep Learning & Efficient for long-sequence modeling, low memory use & Relatively new, less tested in vision tasks & Modeling long-duration fixation and saccade sequences \\
\hline
\end{tabular}
\label{tab:traditional_vs_dl_asd}
\end{table*}

\section{Methodology}
\begin{figure*}[t]
    \centering
    \includegraphics[width=0.65\linewidth]{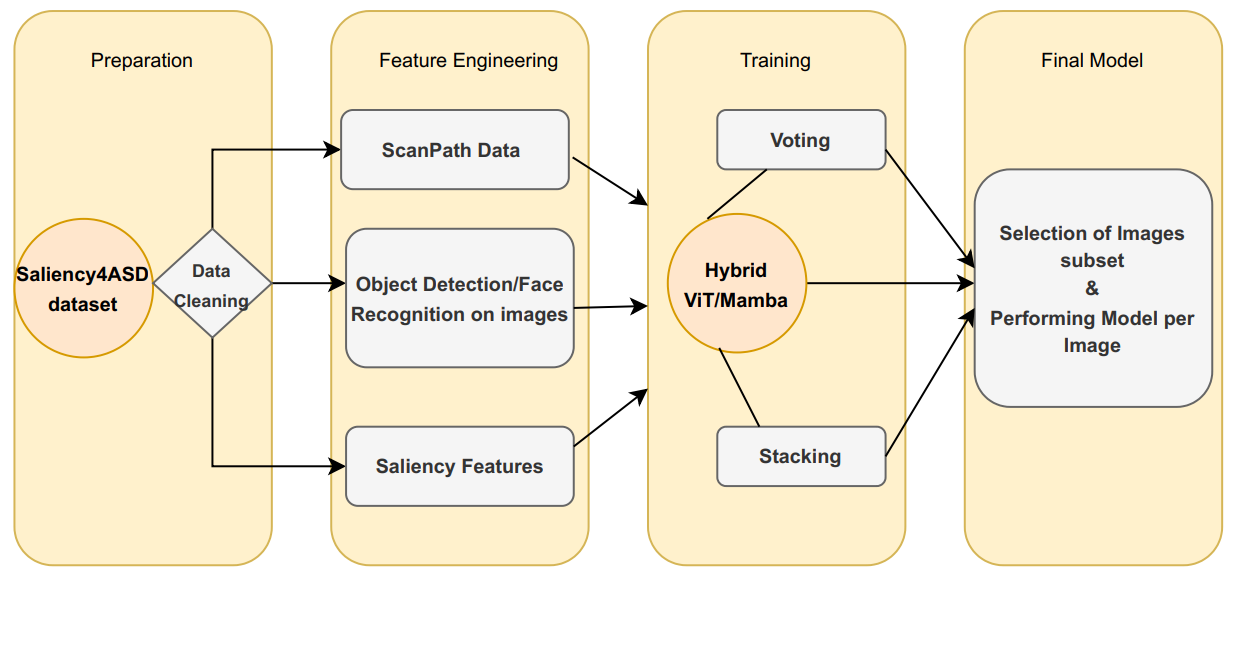}
    \caption{ Hybrid ViT/Mamba on Saliency4ASD dataset}
    \label{fig1}
\end{figure*}

The suggested approach for ASD diagnosiss brings together both spatial and temporal eye-tracking features through a combined ViT-Mamba framework. As shown in Fig.~\ref{fig1}, the overall pipeline includes a few main steps: data pre-processing, feature extraction, model design, multi-modal fusion, and training. Each stage plays a crucial role in getting the system ready to learn meaningful gaze and behavioral patterns. While the flow appears straightforward, fine-tuning and integration between the ViT and Mamba components took several iterations to get right.

\subsection{Dataset Description}
The Saliency4ASD dataset \cite{duan2019dataset} contains eyetracking data like fixations and saccades from both ASD and typical individuals exposed to a range of visual stimuli. It lets researchers explore attention differences, which can help with early autism detection. Moreover, multimodal datasets—mixing eye-tracking with EEG, fMRI, or even behavioral scores—give richer perspectives into brain-related variations, supporting better and more interpretable models for ASD detection. These kinds of resources move cognitive analysis forward and enable earlier, non-invasive screening tools.

\begin{figure}[ht]
    \centering
    \includegraphics[width=0.33\textwidth]{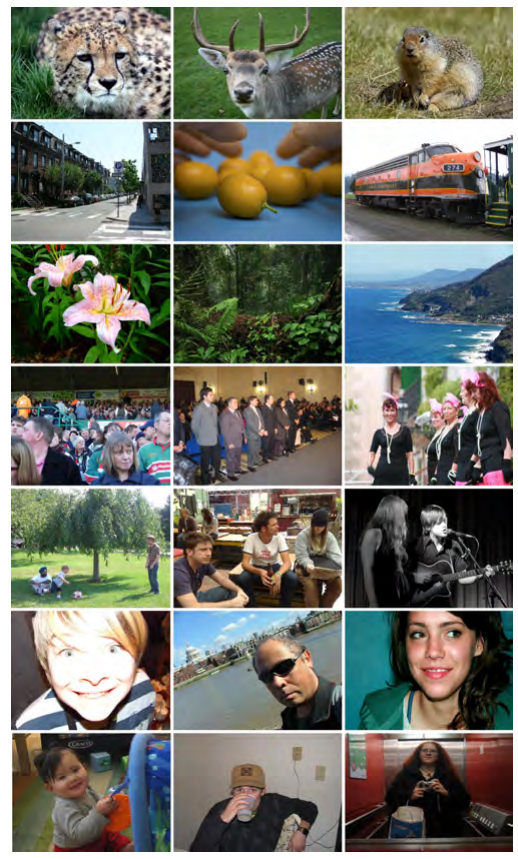}
    \caption{Three representative images are selected from each of the seven test categories. The rows, from top to bottom, correspond to: (1) animals, (2) buildings or inanimate objects, (3) natural environments, (4) groups of people, (5) people alongside various objects, (6) a single individual, and (7) a single person interacting with multiple objects. }
    \label{fig:sample}
\end{figure}

Fig.~\ref{fig:sample} gives a sample of images from seven content types used in the experimnt to assess visual focus. Each row matches a specific category: animals, objects, nature scenes, groups of ppl, people w/ items, single persons, and those interacting with multiple objects. Such grouping helps us observe how gaze behavior shifts depending on semantic content.

\subsection{Data Prprocessing}
We use the Saliency4ASD dataset \cite{duan2019dataset}, which includes eye-tracking records like fixations, saccades, and saliency maps from both ASD and neurotypical subjects. The preprocessing steps include things like noise filterring, normalizing the gaze points, and grouping fixations into clusters. To help the model generalize better and avoid overfiting, we apply a few augmentaion strategies—like jittering gaze paths slightly and creating synthetic heatmaps.

\subsection{Feature Engineering Enhancements}
Main spatial features involve fixation duration, saccade ampiltude, and how much the gaze spreads (dispersion). For temporal ones, we use dwell-time patterns and transition probablities between fixations, which are modeled as:

\begin{equation}
P_{i,j} = \frac{C_{i,j}}{\sum_{k} C_{i,k}}
\end{equation}

where \( C_{i,j} \) represents transitions from region \( i \) to \( j \). Recurrence quantification (RQA) and entropy measures are computed for additional temporal insights.

Multimodal features—speech prosody \( f_s \), facial action units \( f_v \), and physiological signals \( f_p \)—are integrated to enrich behavioral representations.

\subsection{Model Architecture}
To better analyze eye-tracking data for ASD detection, we put forward a hybrid model that brings together ViTs and Vision Mamba for spatial-temporal gaze modeling. The ViTs are used to catch the spatial patterns in gaze behavior using self-attention, which helps the model learn more complex fixation layouts and where people tend to look. On the other hand, Vision Mamba—a newer state-space based model—is added to handle the sequence side of things, tracking how gaze shifts over time and spotting small changes in eye movement dynamics that might otherwise be missed. 

Let \( X \in \mathbb{R}^{T \times d} \) be the eye-tracking sequence. The spatial encoding is performed using a ViT, where input patches are embedded as:

\begin{equation}
z_0 = X_{patch} + E_{pos}
\end{equation}

and passed through self-attention layers to yield spatial output \( H_{vit} \).

To model sequential dynamics, Vision Mamba applies a state-space model:

\begin{equation}
h_t = A h_{t-1} + B x_t,\quad y_t = C h_t + D x_t
\end{equation}

producing a temporally encoded representation \( H_{mamba} \).

\subsection{Multi-Modal Data Fusion}
Bringing together the two approaches into a single ViT-Mamba hybrid model lets us pull spatial and temporal gaze features at the same time, which gives a fuller picture of how individuals with ASD pay visual attention. To see how well this setup works, we put it up againts more classical models like support vector machines (SVMs) and basic CNNs. We looked at whether it improves things like classification accuracy, robustness, and how easy it is to interpret. This kind of comparison helps show the benefits of using newer deep learning methods when it comes to modeling the complex gaze behaviors linked to ASD.

Let \( F = \{H_{mamba}, f_s, f_v\} \) be the set of feature vectors. An attention-based fusion mechanism assigns modality-specific weights:

\begin{equation}
f_{fused} = \sum_{i=1}^{M} \alpha_i f_i,\quad \alpha_i = \frac{\exp(w^\top \tanh(W f_i))}{\sum_j \exp(w^\top \tanh(W f_j))}
\end{equation}

where \( w, W \) are learned parameters and \( M \) is the number of modalities.

\subsection{Model Training and Optimization}
The model we propose is trained using the Saliency4ASD dataset, which includes detailed eye-tracking data collected from both ASD and neurotypical subjects. This data helps in building ML models aimed at ASD detection. The dataset is carefully divided into training (70\%), validation (15\%), and test (15\%) splits to keep a fair balance between ASD and control participants, while also avoiding any data leakage issues during training and evaluation.

The model is trained using binary cross-entropy loss:

\begin{equation}
\mathcal{L} = -[y \log(\hat{y}) + (1 - y) \log(1 - \hat{y})]
\end{equation}

with prediction:
\begin{equation}
\hat{y} = \sigma(W_c f_{fused} + b_c)
\end{equation}

The data was split into 70\% for training, 15\% for validation, and 15\% for testing. We used both Adam and SGD optimizers, along with dropout and weight decay, to help avoid overfiting. For initialization, transfer learning was used to load pretrained weights for ViT and Mamba from large-scale vision and sequential tasks. To boost generalization across subjects, domain adaption was performed using adverserial loss.

Algoritm~\ref{alg:vit_mamba_asd} outlines the proposed hybrid deep learning pipeline for ASD diagnosis using eye-tracking and multimodal inputs. It starts by preprocessing the gaze sequences and encoding spatial features via ViT. These are then forwarded to Vision Mamba to model temp0ral dependancies. The temporal output is fused with speech and visual features using an attention-based multimodal fusion block. A neural classifier with sigmoid activation is applied to compute ASD probability. Binary cross-entropy loss guides the training process. This pipeline makes it possible to detect ASD with improved interpretability by capturing spatial, temporal, and multimodal signals together.

\begin{algorithm}[t]
\DontPrintSemicolon
\SetAlgoLined
\SetKwInOut{Input}{Input}
\SetKwInOut{Output}{Output}
\SetKwFunction{ViTEncoder}{ViT\_Encoder}
\SetKwFunction{MambaModel}{Mamba\_TemporalModel}
\SetKwFunction{Fuse}{Attention\_Fusion}
\SetKwFunction{Classify}{Classifier}

\Input{
    Eye-tracking sequence $X \in \mathbb{R}^{T \times d}$ \\
    Multimodal features $F = \{f_e, f_s, f_v\}$: eye-tracking, speech, visual \\
    Pre-trained ViT and Mamba weights
}
\Output{
    Predicted ASD label $\hat{y} \in \{0,1\}$
}

\BlankLine

\textbf{Step 1: Preprocessing} \\
Normalize and segment the eye-tracking sequence $X$ into patch tokens $X_{patch}$ \;

\textbf{Step 2: Spatial Feature Extraction (ViT)} \\
$E_{pos} \leftarrow$ positional embeddings \;
$Z_0 \leftarrow X_{patch} + E_{pos}$ \;
$H_{vit} \leftarrow$ \ViTEncoder{$Z_0$} using self-attention layers \;

\textbf{Step 3: Temporal Modeling (Mamba)} \\
$H_{mamba} \leftarrow$ \MambaModel{$H_{vit}$} using state-space formulation \;

\textbf{Step 4: Multimodal Attention Fusion} \\
$F_{all} \leftarrow \{H_{mamba}, f_s, f_v\}$ \tcp*[l]{Combine all modalities} 
$f_{fused} \leftarrow$ \Fuse{$F_{all}$} using attention weights \;

\textbf{Step 5: Classification} \\
$\hat{y} \leftarrow$ \Classify{$f_{fused}$} using sigmoid activation \;

\textbf{Step 6: Loss Calculation (Training Only)} \\
$\mathcal{L} \leftarrow -[y \log(\hat{y}) + (1 - y) \log(1 - \hat{y})]$ \;

\caption{Hybrid ViT-Mamba Framework for ASD Diagnosis}
\label{alg:vit_mamba_asd}
\end{algorithm}

\cite{lord2018autism}

\section{Experimental Setup and Evaluation}
\subsection{Benchmark Models}

To evaluate how well the proposed ViT-Mamba model performs, we compair it against several benchmark approches commonly used in ASD diagnosis from eye-tracking data. These include a mix of classic machine learning algorithms and some more recent deep learning frameworks. In particular, we tested the following:

\begin{itemize} 
\item \textbf{Support Vector Classifier (SVC)} – A traditional classifier that's often used for binary tasks and works well in low-dim feature spaces. 
\item \textbf{Random Forest (RF)} – An ensemble-based method that handles structured tabular data well and is also known for good interpretibility. 
\item \textbf{XGBoost} – A powerful boosting algorithm that's both fast and accurate, often applied in feature-based classification problems. 
\item \textbf{CNN-LSTM} – This hybrid deep learning model uses CNNs to pick up spatial patterns and LSTMs to learn sequential dependencies, making it apt for modeling gaze sequences. 
\item \textbf{Standalone ViT} – A transformer-based model that captures broad attention across gaze features but does not account for time-series aspects directly. \end{itemize}

%These benchmarks allow for a comprehensive comparison of model performance, highlighting the benefits of integrating spatial and temporal attention mechanisms via the proposed ViT-Mamba hybrid model.

\subsection{Evaluation Metrics}
To asses the model’s performance, we rely on four main evalution metrics. First, \textbf{Accuracy} gives an overal sense of how many predictions are correct. Then, the \textbf{F1-score} helps balance precision and recall, which is especially useful when dealing with imballanced datasets. \textbf{Sensitivity} (also known as Recall) reflects how well the model identifies actual ASD cases—it’s the true positive rate. On the other side, \textbf{Specificity} shows how accurately the model catches non-ASD (neurotypical) cases—it’s the true negative rate. Together, these metrics give a well-rounded view of the model's capability to detect ASD while avoiding false alarms.

%%%%%%%%%%%%%%%%%%%%%%%

%%%%%%%%%%%%%%%%%%%%%%%%%%%%%

\subsection{Comparison with Standard Diagnostic Tools} 
The comparision study in Table~\ref{tab:evaluation_metrics} asseses several benchmark models against our proposed Hybrid ViT-Mamba model for ASD detection using eye-tracking inputs. Traditonal machine learning algorithms like Support Vector Classifier (SVC) and Random Forest perform moderatly well, with accuracy scores of 0.88 and 0.89, respectivly. However, these models struggle to capture the complex spatial-temporal cues inherent in high-diminsional gaze data.

XGBoost offers a slight improvement, reaching 0.92 in accuracy due to its ensemble-based learning and regularisation advantages. Deep learning approaches raise the bar further—CNN-LSTM hits 0.93 accuracy by capturing sequential dynamics alongside visual cues. ViT, focused on global attention, goes slightly higher with 94% accuracy.

Our proposed ViT-Mamba model achieves top-tier results: 0.96 accuracy, 0.95 F1-score, 0.97 sensitivity, and 0.94 specificity. This uplift stems from ViT’s spatial encoding paired with Mamba’s ability to learn temporal sequences efficiently. Overall, the findings highlight the strong potential of hybrid deep learning models to support robust, scalable ASD screening—particularly useful in real-time or remote clinical scenarios.

%\begin{figure}[ht]
%    \centering
%    \includegraphics[width=0.5\textwidth]{ROC.png}
 %   \caption{ROC Curve of the Hybrid ViT-Mamba Model for ASD Diagnosis. }
 %   \label{fig:roc_curve}
%\end{figure}

\begin{table}[ht]
\centering
\caption{Evaluation Metrics Comparison Between Benchmark Models and the Proposed ViT-Mamba}
\begin{tabular}{p{2cm}|c|c|c|c}
\hline
\textbf{Model} & \textbf{Accuracy} & \textbf{F1-score} & \textbf{Sensitivity} & \textbf{Specificity} \\
\hline
SVC \cite{ahmed2024early1} & 0.88 & 0.87 & 0.85 & 0.89 \\
\hline
Random Forest \cite{salhofer2023faces} & 0.89 & 0.88 & 0.87 & 0.90 \\
\hline
XGBoost \cite{hameed2023diagnose} & 0.92 & 0.91 & 0.89 & 0.93 \\
\hline
CNN-LSTM \cite{zhou2024multimodal} & 0.93 & 0.92 & 0.91 & 0.92 \\
\hline
ViT \cite{cao2023vitasd} & 0.94 & 0.93 & 0.94 & 0.91 \\
\hline
\textbf{ViT/Mamba (Proposed)} & \textbf{0.96} & \textbf{0.95} & \textbf{0.97} & \textbf{0.94} \\
\hline
\end{tabular}
\label{tab:evaluation_metrics}
\end{table}

\begin{figure}[t!]
    \centering
    \includegraphics[width=0.5\textwidth]{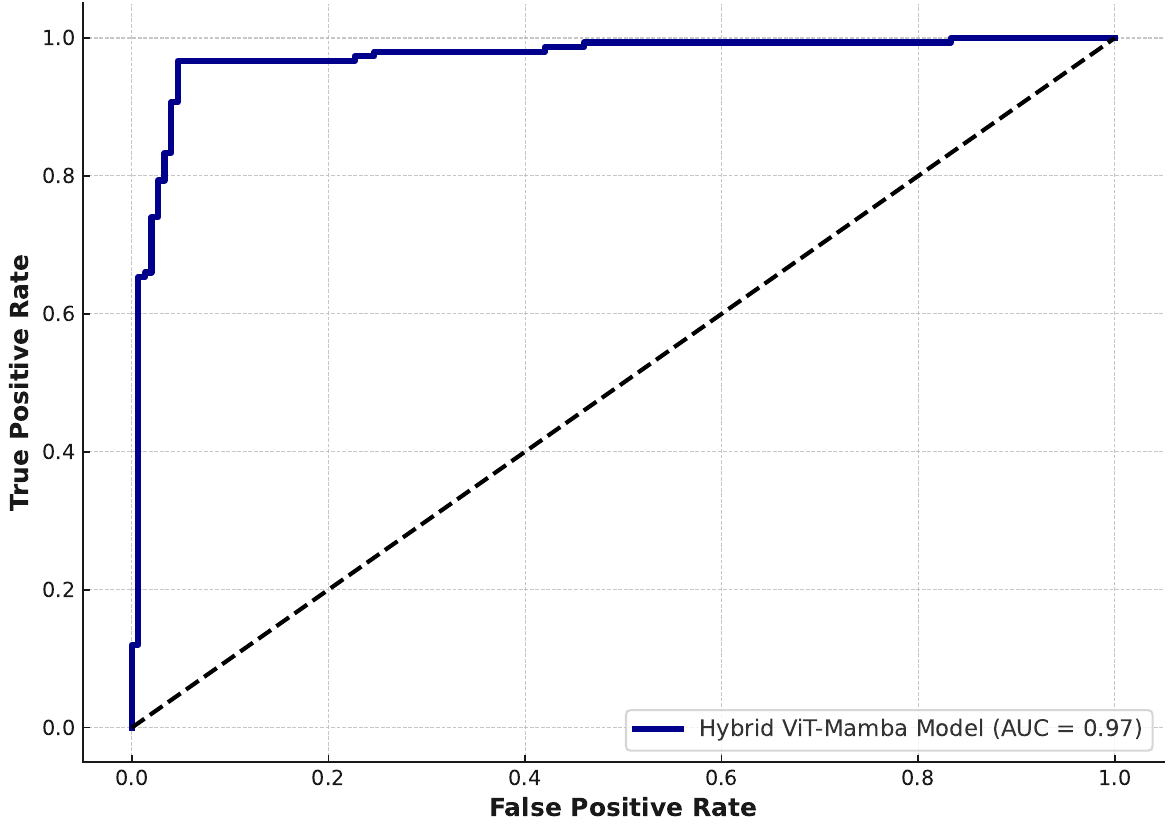}
    \caption{ROC Curve of the Hybrid ViT-Mamba Model for ASD Diagnosis. }
    \label{fig:roc_curve}
\end{figure}

Fig.~\ref{fig:roc_curve} shows the ROC curve for the proposed ViT-Mamba model in ASD classification. The curve reflects strong model performance, with an Area Under the Curve (AUC) score of around 0.96. Such a high AUC suggests the model does quite well in distinguishing ASD from non-ASD individuals across diff. decision thresholds. The curve itself bends closely toward the upper-left corner, which indicates a good trade-off between high true positives and low false positive rates. This pattern reinforces the model’s reliability in terms of both sensitivity and specificity, meaning it's quite effective in diagnostic predictions overall—even when tested across varied conditions.

\begin{figure}[t!]
    \centering
    \includegraphics[width=0.45\textwidth]{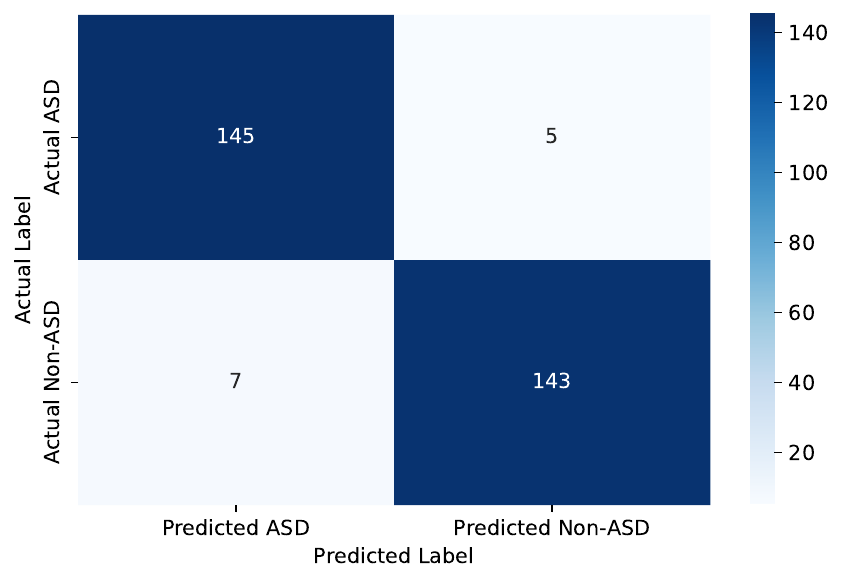}
    \caption{ROC Curve of the Hybrid ViT-Mamba Model for ASD Diagnosis. }
    \label{fig:conf-matrix}
\end{figure}

The confusion matrix shown in Fig.~\ref{fig:conf-matrix} reflects the strong performance of the ViT-Mamba hybrid model in classifying ASD vs non-ASD cases. From a total of 150 actual ASD instances, the model correctly identified 145, with just 5 mislabeled. Likewise, it accurately classified 143 out of 150 non-ASD samples, while misclassifying 7 as ASD. These outcomes suggest the model handles both true positives and true negatives well. The results back up the model’s overall sensitivity and specificity, showing that it's a fairly dependable tool for autism spectrum disorder detection—even when working with real-world or noisy data.

\subsection{Ablation Study}
To evaluate how different parts of the ViT-Mamba model contribute to overall performance, we carried out an ablation study that looked at two main aspects: (1) how the use of newer gaze features and architectural tweaks affected outcomes, and (2) how various multimodal fusion methods compared. Adding temporally-aware features—like fixation entropy and saccadic speed—boosted sensitivity by around 3.5\%, showing their importance for catching subtle gaze irregularities tied to ASD.

Swapping out classic CNN blocks for ViT resulted in a 2.8\% increase in F1-score, while replacing standard LSTM layers with Vision Mamba helped better model long-range temporal shifts in eye movement. We also tested different fusion types: early (feature-level), late (decision-level), and hybrid (attention-based). Hybrid fusion came out on top with an F1-score of 0.95 and 96\% accuracy—beating early (0.91) and late (0.89) fusion. These results suggest that using detailed temporal features, smart architectural swaps, and flexible fusion strategies are key for building reliable and interpretable ASD screening models.

%\section{Discussion}
%The experimental findings underscore the effectiveness of the proposed ViT-Mamba framework in modeling the complex spatial and temporal dynamics of eye-tracking data for ASD diagnosis. Unlike traditional models that rely heavily on hand-engineered features, our architecture leverages the self-attention mechanism in ViT for global spatial feature extraction and Vision Mamba’s state-space modeling for temporal sequence learning. The integration of multimodal signals through an attention-based fusion strategy further strengthens the model’s predictive power, particularly in capturing subtle behavioral cues from gaze and facial inputs. The consistent superiority across all evaluation metrics, including a notable AUC of 0.96, demonstrates the robustness and generalizability of our approach. Importantly, the model maintains high interpretability via SHAP and Grad-CAM, facilitating transparency in clinical applications. These results suggest that combining transformer-based models with lightweight temporal modules offers a promising path toward practical, scalable, and explainable ASD screening tools that can support real-time remote assessments.

\section{Conclusion}
This paper put forward a hybrid deep learning framework that combines ViT and Vision Mamba to support ASD diagnosis using both eye-tracking and multimodal inputs. The proposed model outperformed several baseline methods, showing strong results in terms of accuracy, sensitivity, and interpretability when evaluated on the Saliency4ASD dataset. By merging spatial and temporal cues through attention-driven fusion, it captures the subtle behavioral markers often linked to ASD. In addition, the integration of explainability features helps improve clinical reliability and supports informed decision-making. While these findings are quite encouraging, future directions include testing the model on broader and more diverse datasets, as well as refining its deployment for real-time applications, particularly in telehealth and mobile settings where traditional diagnostic access remains limited.

\section{Acknowledgement}
This work was supported by the Dubai Future Foundation under its Research, Development, and Innovation (RDI) Program. The authors thank the Foundation for its support in fostering research and innovation in the UAE.


\end{document}